\documentclass{article}
\usepackage{ijcai22}

\usepackage{times}
\usepackage{soul}
\usepackage{url}
\usepackage[hidelinks]{hyperref}
\usepackage[utf8]{inputenc}
\usepackage[small]{caption}
\usepackage{graphicx}
\usepackage{amsmath}
\usepackage{amsthm}
\usepackage{amssymb}
\usepackage{booktabs}
\usepackage{algorithm}
\usepackage{algorithmic}
\usepackage{tikz}
\usetikzlibrary{arrows,shapes}
\usepackage[textwidth=15mm,textsize=footnotesize, disable]{todonotes}

\newtheorem{example}{Example}

\newtheorem{definition}{Definition}

\title{On Interactive Explanations as Non-Monotonic Reasoning}

\newcommand{\orcidID}[1]{}
\author{
Guilherme Paulino-Passos\orcidID{0000-0003-3089-1660}$^1$
\and
Francesca Toni\orcidID{0000-0001-8194-1459}$^1$
\affiliations
$^1$Imperial College London, Department of Computing\\
\emails
\{g.passos18, f.toni\}@imperial.ac.uk
}

\usepackage[normalem]{ulem} % for \sout

\newcommand{\gpp}[2][]{\todo[author=Guilherme,caption={}, #1]{#2}}
\newcommand{\gppnew}[1]{\textcolor{olive}{#1}}
\newcommand{\gppold}[1]{\textcolor{olive}{\sout{#1}}}

\newcommand{\FTdel}[1]{}

\makeatletter
\if@todonotes@disabled
\renewcommand{\gppnew}[1]{#1}

\renewcommand{\gppold}[1]{} % uncomment to hide the "old" text
\renewcommand{\FTdel}[1]{}
\else
\fi
\makeatother

\makeatletter
\if@todonotes@disabled
\else
\fi
\makeatother

\newif\ifincludeincoherent
\includeincoherenttrue

\newif\ifincludeproofs
\includeproofsfalse

\newif\ifincludeNN              %n earest neighbours
\includeNNfalse

\renewcommand{\phi}{\varphi} % I prefer this phi :)

\tikzset{attack/.style={-latex}}

\mathchardef\mhyphen="2D

\newcommand{\inputset}{\ensuremath{X}}
\newcommand{\outputset}{\ensuremath{Y}}
\newcommand{\classifier}{\mathbb{C}}

\newcommand{\explainer}{\ensuremath{\mathbb{E}}}
\newcommand{\explainerindiv}{\mathbb{E}_{\bullet}} % bullet for "point"

\newcommand{\explanation}{\ensuremath{E}}
\newcommand{\explanations}{\mathcal{E}} % for the _set_ of
\newcommand{\entail}{\ensuremath{\models}} % entailment, at explanation-level
\newcommand{\entails}{\entail}
\newcommand{\entailone}{\ensuremath{\models}} % entailment from one explanation
\newcommand{\deriveio}{\ensuremath{\vdash}} % derive at instance, IO-level
\newcommand{\deriveexp}{\ensuremath{\vdash_{e}}} % derive at explanation-level
\newcommand{\derivegen}{\ensuremath{\vdash'}} % derive generic, used to define properties
\newcommand{\setgen}{\ensuremath{S}} % generic set

\newcommand{\seqsetdef}[1]{\ensuremath{\bigcup_{i \in \naturalnumb} #1^i}}
\newcommand{\seqset}[1]{\ensuremath{Seq(#1)}}
\newcommand{\naturalnumb}{\mathbb{N}}
\newcommand{\realnumb}{\mathbb{R}}
\newcommand{\seq}[2][n]{\ensuremath{(#2_i)_{i \in [#1]}}}
\newcommand{\seqio}[1][n]{\ensuremath{(x_i,y_i)_{i \in [#1]}}}

\makeatletter
\if@todonotes@disabled
  \usepackage{named} %

  \newcommand{\citet}[2][]{\citeauthor{#2}~\shortcite{#2}}

  \newcommand{\citep}[2][]{\cite{#2}}

\else
  \usepackage{natbib}
\fi
\makeatother
\begin{document}

\maketitle

\begin{abstract}
  
  Recent work shows issues of consistency with explanations, with methods generating local explanations that seem reasonable instance-wise, but that are inconsistent across instances. This suggests not only that instance-wise explanations can be unreliable, but mainly that, when interacting with a system via multiple inputs, a user may actually lose confidence in the system.
  To better analyse this issue, in this work we treat explanations as objects that can be subject to reasoning and present a formal model of the interactive scenario between user and system, via sequences of inputs, outputs, and explanations. We argue that explanations can be thought of as committing to some model behaviour (even if only \textit{prima facie}), suggesting a form of entailment, which, we argue, should be thought of as non-monotonic. This allows: 1) to solve some considered inconsistencies in explanation, such as via a specificity relation; 2) to consider properties from the non-monotonic reasoning literature and discuss their desirability, gaining more insight on the interactive explanation scenario.
\end{abstract}

\section{Introduction}
\label{sec:intro}
A growing body of research is dedicated to provide (post-hoc) explanations to AI models, in particular machine learning systems. However, there is no standardised formal definition of explanation, and what %
properties explanations should satisfy, with different explanation methods having different goals, %
often implicit. %

Issues of consistency have been raised in recent work. \citet{camburu-etal-2020-make} show how inconsistent textual explanations can be generated for a natural language inference task. \citet{Merrer2020RemoteEF} argue that local explanations for a single prediction are always subject to manipulability by hiding how a protected feature was used, but present, as a method of detecting this manipulation, that an auditor queries the system multiple times and detects whether there is an inconsistency in the returned explanations.
Thus, lack of consistency may not only cause a user to wrongly expect %
some system behaviour, but may also cause loss of user's confidence in the system overall, or imply that the explanations are not faithful, or even malicious.
Therefore it is increasingly important to discuss the meaning of consistency in the case of local explanations, an issue that can occur when querying a system multiple times, with different inputs.%

While the idea of consistency may first evoke classical logic, a well-established tradition in logic-based AI argues that in many applications the form of reasoning more natural for humans it not classical logic, but different models of reasoning which are \emph{non-monotonic}, meaning that, in the presence of extra premises, previous conclusions may become unwarranted \citep{generalpatterns,DBLP:journals/ai/KrausLM90}. Indeed, a line of work in psychology empirically evaluates inferences humans make in a task with conditional arguments. Understanding explanations as presenting conditional arguments for expected outputs given some inputs would, then, lead one to model explanations as non-monotonic reasoning (NMR).

\gppnew{This is precisely what we do here,} proposing an analysis of reasoning with explanations for an interactive scenario. We think of explanations as objects that 
can be reasoned upon.
Thus we model explanations abstractly and define which inferences we can make from them, %
while also considering which reasoning properties they may satisfy
. \gppnew{Our framework allows one to define analogous of traditional properties of NMR, and we hypothesise which properties might be expected or desirable.}

\section{The interactive explanation process}
\label{sec:interactive-process}

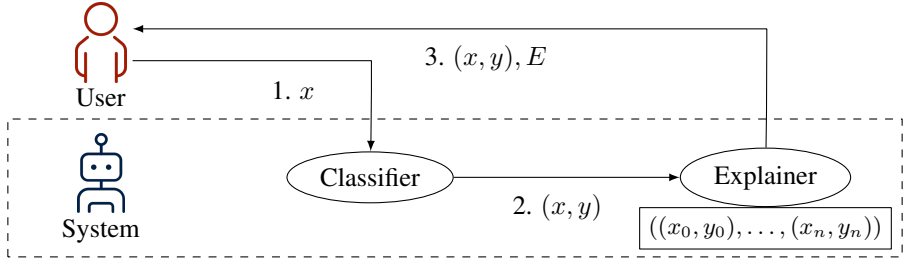
\begin{figure*}[t]
  \centering
  \begin{tikzpicture}[>=latex,line join=bevel,]
    \begin{scope}
      \pgfsetstrokecolor{black}
      \pgfsetdash{{3pt}{3pt}}{0pt}
      \definecolor{strokecol}{rgb}{0.0,0.0,0.0};
      \pgfsetstrokecolor{strokecol}
      \draw [dashed] (8.0bp,8.0bp) -- (8.0bp,60.0bp) -- (345.0bp,60.0bp) -- (345.0bp,8.0bp) -- cycle;
    \end{scope}
    \node (robot) at (43.0bp,34.0bp) [draw,draw=none,align=center] {\includegraphics*[height=3em]{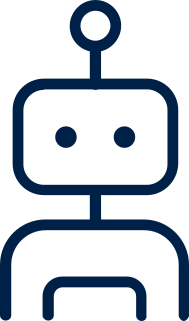}\\[-0.2em]System};
    \node (class) at (144.5bp,38.0bp) [draw,ellipse] {Classifier};
    \node (expl) at (293.5bp,38.0bp) [draw,ellipse] {Explainer} ;
    \node (history) [draw,rectangle,below=0em of expl] {\footnotesize{$((x_0, y_0), \dots, (x_n, y_n))$}};    
    \node (human) at (43.0bp,88.0bp) [draw,draw=none,align=center] {\includegraphics*[height=3em]{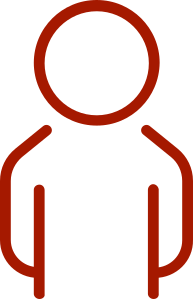}\\[-0.2em]User\vspace{-0.75em}};
    \draw [->] (class) ..controls (183.26bp,38.0bp) and (245.96bp,38.0bp)  .. (expl);
    \draw (215.11bp,26.8bp) node {2. $(x,y)$};
    \draw [->] (55.0bp,82.0bp) ..controls (100.28bp,82.0bp) and (145.0bp,82.0bp)  .. (145.0bp,82.0bp) .. controls (145.0bp,82.0bp) and (145.0bp,59.103bp) .. (class);
    \draw (115.48bp,70.8bp) node {1. $x$};
    \draw [->] (expl) ..controls (294.0bp,65.384bp) and (294.0bp,94.0bp)  .. (294.0bp,94.0bp) .. controls (294.0bp,94.0bp) and (80.339bp,94.0bp)  .. (55.0bp,94.0bp);
    \draw (188.14bp,82.8bp) node {3. $(x,y), \explanation$};
  \end{tikzpicture}

  \caption{%
    Overview of the \emph{interactive explanation} process. 1. The user queries the system with an input $x$%
    , which goes to the classifier%
  . 2. The classifier produces %
  output $y%
  $, and sends the pair $(x,y)$ to the explainer. 3. The explainer produces an explanation $E$ %
  for %
  $(x,y)$, using information %
  from the history %
  of %
  inputs/outputs %
  $((x_0, y_0), \dots, (x_n, y_n))$%
  , and sends %
  $(x,y)$ and $E$ back to the user%
  , who
  may then stop %
  or %
  further query %
  the system.}
\label{fig:inter-process}
\end{figure*}

Consider the scenario of a %
user or auditor evaluating the behaviour of a model. They may ask for the output for a specific input, and ask as well for an explanation. While much of this literature treats this as %
a one-off task
, a single explanation may be insufficient. The user could keep exploring model behaviour by querying for more outputs (with the same or different inputs) and explanations of the model. We thus consider an interactive process (as overviewed in Fig.~\ref{fig:inter-process}), where the user queries for an output and explanation thereof, given an input, the %
system returns %
them, and %
then the user may query again. %
We see the AI system as including a \emph{classifier}\footnote{This choice is dictated by the focus, in the XAI literature, on explaining outputs of classifiers. However, our conceptual understanding of the interactive explanation process is applicable to any model.} and an \emph{explainer}, %
with both
considered black-boxes by our conceptual %
model.

We %
assume that the explanation method (although not the classifier) can keep track of the \textit{history} of %
inputs it received and their outputs. This allows %
supporting two scenarios: i) of an explainer that tries to improve its explanations by knowing what it explained before%
; and ii) of a malicious explainer that%
, trying to manipulate or mislead the user and to avoid being detected, keeps track of what was explained before. 
Histories give a  snapshot of the process by finite sequences of %
inputs and their outputs%
.
Note that
we assume the explainer to be a function, thus ignoring randomness of the explanation method itself. This assumption implies that no information about the computed explanations needs to be stored in the history, as awareness of earlier inputs and outputs %
already gives all information need to know what explanations were also previously returned%
.\footnote{
  Our assumption here simplifies our modelling, especially Fig.~\ref{fig:diagram} (discussed later). In case it is not deterministic, such as in method that includes sampling, %
past explanations should be %
included directly in the history. We leave this extension of our model to future study.}

\begin{example}
  \label{ex:setup}
  For concreteness, let us consider a scenario where inputs are drawn from $\inputset{} = \realnumb^2$,
  outputs %
  from $\outputset = \{0,1\}$, and %
  explanations ($\explanations$) %
  are decision sets \citep{DBLP:conf/kdd/LakkarajuBL16}, that is, sets of rules, each of the form $s \rightarrow y = c$, where $s$ is an itemset (a conjunction of predicates of the form (feature, operator, value), such as $%
  f > 1$), and $c \in \outputset$.\footnote{However, as opposed to the original definition in \citep{DBLP:conf/kdd/LakkarajuBL16}, we %
  assume no default class% 
  and no tie-breaking function.}
  
  A possible scenario of interaction is as follows. The user queries for $x_0 = (5,0)$, corresponding to feature assignments $f=5,g=0$; %
  the classifier returns $y_0 = 1$, and the explanation given is $\explanation_0 = \{%
  f > 0 \rightarrow y = 1\}$%
  . The user decides to investigate further and tries $x_1 = (20,5)$. Even though this is covered by the previous rule, which would commit to the output $y_1=1$, suppose the output is $y_1 = 0$ and the explanation %
  is $\explanation_1 = \{%
  f> 10 \land %
  g > 3 \rightarrow y = 0\}$. How %
  should the user %
  interpret this? Is this an inconsistency in the explanatory process?
\end{example}

\section{Formal modelling}
\label{sec:formal-modelling}

We assume an input set $\inputset$ and an output set $\outputset$ %
as well as a set of possible explanations $\explanations$, that we keep abstract. Regarding notation, for a set $S$, we denote the set of finite sequences of element of $S$ as $\seqset{S}$, i.e., $\seqset{S} = \seqsetdef{S}$ for $S^i$ a sequence of $i$ elements of $S$. Given $n \in \naturalnumb$, we use the notation $[n] = \{m \in \naturalnumb \mid m \leq n\}$. Thus a sequence $(s_0, s_1, \dots, s_n) \in \seqset{S}$ %
can be written as $(s_i)_{i \in [n]}$.

We consider that the system is composed of a classifier ${\classifier: \inputset \rightarrow \outputset}$ and an explanation method ${\explainer: {\seqset{{\inputset\times\outputset}} \rightarrow \seqset{\explanations}}}$, %
mapping from %
a sequence of input-output pairs $(x_i, y_i)_{i \in [n]}$ to a sequence of explanations $(E_i)_{i \in [n]}$, of the same length as the sequence of pairs. Notice that the explainer uses information %
on the entire past of %
inputs-outputs%
, as we discussed in Section \ref{sec:interactive-process}. We can think of this sequence of pairs as a \textit{history}%
.
We %
consider that at each time step $t \in \naturalnumb$, the user queries for an input $x_t \in \inputset$, which receives a classification $\classifier(x_t) = y_t$ and an explanation $\explanation_t$. In this way, the explainer provides an explanation motivated by a specific %
input-output, while considering the history
$((x_i, y_i))_{i \in [t-1]}$. %

An important particular case is when there is a function $\explainerindiv:%
\inputset \times \outputset \rightarrow \explanations$, mapping from %
a single example $(x, y)$ to an explanation $\explanation$. In this case, the explainer function $\explainer$ can be defined %
as applying
$\explainerindiv$ to each element of the sequence: formally, $\explainer(%
((x_i, y_i))_{i \in [n]}) = (\explainerindiv(%
x_i, y_i))_{i \in [n]}$.
In this particular case the history is disregarded when explanations are computed.

The reader may notice that this %
view of the interactive explanation process does not enforce that previously exhibited explanations are kept, that is, that $E_%
i$ is unchanged for all $i \in [t]$ when $x_{t+1}$ is queried. If 
instead we want past explanations to be unretractable, in the sense that the system cannot replace the explanation of %
any previous query, then 
we can enforce the following %
 property:
\begin{definition}[Interaction-stability]
    An explainer \explainer{} is said to be \emph{interaction-stable} whenever, for every sequence of input-output pairs $(x_i,y_i)_{i \in [n]}$ and for every $m <n$,
  if
  $(E_i)_{i \in [n]} = \explainer((x_i, y_i)_{i \in [n]})$ and $(E'_i)_{i \in [m]} = \explainer((x_i, y_i)_{i \in [m]})$ then   $E_i = E'_i$
  for any $i \in [m]$.
\end{definition}
That is, an interaction-stable %
explainer
will always keep the explanation $E_i$ associated to the pair $(x_i, y_i)$, even as the interaction moves on.
It is straightforward to see that an explainer $\explainer$ derived from a function $\explainerindiv{}$ is always interaction-stable.

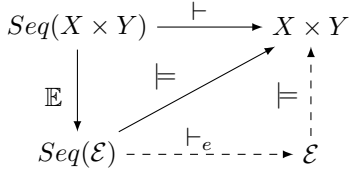
\begin{figure}[t]
  \centering
  \begin{tikzpicture}[>=latex,line join=bevel,]
\begin{scope}
  \pgfsetstrokecolor{black}
  \definecolor{strokecol}{rgb}{1.0,1.0,1.0};
  \pgfsetstrokecolor{strokecol}
  \definecolor{fillcol}{rgb}{1.0,1.0,1.0};
  \pgfsetfillcolor{fillcol}
  \filldraw (0.0bp,0.0bp) -- (0.0bp,79.26bp) -- (128.0bp,79.26bp) -- (128.0bp,0.0bp) -- cycle;
\end{scope}
\begin{scope}
  \pgfsetstrokecolor{black}
  \definecolor{strokecol}{rgb}{1.0,1.0,1.0};
  \pgfsetstrokecolor{strokecol}
  \definecolor{fillcol}{rgb}{1.0,1.0,1.0};
  \pgfsetfillcolor{fillcol}
  \filldraw (0.0bp,0.0bp) -- (0.0bp,79.26bp) -- (128.0bp,79.26bp) -- (128.0bp,0.0bp) -- cycle;
\end{scope}
  \node (A) at (22.0bp,67.26bp) [draw,draw=none] {$\seqset{X \times Y}$};
  \node (B) at (22.0bp,19.26bp) [draw,draw=none] {$\seqset{\explanations}$};
  \node (D) at (110.0bp,19.26bp) [draw,draw=none] {$\explanations$};
  \node (C) at (110.0bp,67.26bp) [draw,draw=none] {$X \times Y$};
  \draw [->] (A) -- (B);
  \definecolor{strokecol}{rgb}{0.0,0.0,0.0};
  \pgfsetstrokecolor{strokecol}
  \draw (13.0bp,40.76bp) node {$\explainer$};
  \draw (16.0bp,52.554bp) node {$$};
  \draw (16.0bp,29.026bp) node {$$};
  \draw [->, dashed] (D) -- (C);
  \draw (101.0bp,41.51bp) node {$\entailone$};
  \draw (104.0bp,29.026bp) node {$$};
  \draw (116.0bp,53.983bp) node {$$};
  \draw [->] (A) -- (C);
  \draw (68.0bp,73.26bp) node {$\deriveio$};
  \draw (50.206bp,61.26bp) node {$$};
  \draw (85.946bp,61.26bp) node {$$};
  \draw [->, dashed] (B) -- (D);
  \draw (68.0bp,25.26bp) node {$\deriveexp$};
  \draw (31.781bp,13.26bp) node {$$};
  \draw (100.33bp,13.26bp) node {$$};
  \draw [->] (B) -- (C);
  \draw (55.0bp,48.5599bp) node {$\entail$};
  \draw (19.506bp,10.344bp) node {$$};
  \draw (100.24bp,54.135bp) node {$$};
\end{tikzpicture}

%%% Local Variables:
%%% mode: latex
%%% TeX-master: "xlokr2022"
%%% End:
 %
  \vspace{-1em}
  \caption{Diagram illustrating the relations between sequences of %
    input-output pairs, sequences of explanations, single %
    pairs and single explanations. This diagram does not always commute (indicated by dashed lines) since for a sequence of explanations \deriveio-entailing an %
    input-output pair, there may not exist a single explanation \deriveexp-entailed by this sequence which \entailone-entails this %
    pair.}
  \label{fig:diagram}
\end{figure}

We now turn to the modelling of inference. We assume that a sequence of explanations
$(\explanation_i)_{i \in [n]}$ ``commits'' to some model behaviour. We model this by an entailment relation $\entails$ between $\seqset{\explanations}%
$ and $X \times Y$, in such a way that $(\explanation_i)_{i \in [n]}  \entails (x,y)$ means that $(\explanation_i)_{i \in [n]}$
``commits'' to the outcome $y$, given the input $x$. 
 We will abuse notation %
 and define $\explanation \entails (x,y)$ to mean $(\explanation) \entails (x,y)$ (for $(E)$ the sequence with just one  explanation, $E$).
 This entailment relation we keep abstract and application-dependent. What is important is that it captures how a user would interpret the explanation %
 or plausible inferences
 therefrom, including as regards the input-output being explained. One example %
is explanations as sufficient reasons: any sufficient reason is exactly a rule that guarantees the %
output for a part of the input space, including the given input.
\gppnew{An important particular case of entailment is when it does not depends on the order of the elements of the sequence. In this case, a set-based representation would be enough, and it is in this sense that sequences generalise sets.}

\addtocounter{example}{-1}
\begin{example}[continued]
  \label{ex:nonmono}
  Assume the entailment relation $\models$ from sequences of explanations $(\explanation_i)_{i \in [n]}$ is to be interpreted as %
  classification from the %
  union %
  $\bigcup_{i \in [n]} \explanation_i$
  of all explanations, corresponding to a naive interpretation of decision sets without tie-breaks.
  Then, $x_1$ is covered by both $\explanation_0$ and $\explanation_1$, for different classifications, which the user %
  may take to be an inconsistency, since no input can have two different outputs.
  A different intended interpretation of the sequence of rules is possible. Define a rule to be more specific than another whenever every input $x$ that satisfies the itemset of the first rule also satisfies the itemset of the second rule. Now assume that only most specific rules are applicable. For this example, $\explanation_1$ is more specific than $\explanation_0$, %
  solving the
  inconsistency issue. This interpretation is \textit{non-monotonic} since, while $(\explanation_0) \entails (x_1, 0)$, we have $(\explanation_0, \explanation_1) \entails (x_1, 1)$, while $(\explanation_0, \explanation_1) \not\entails (x_1, 0)$. That is, adding a new explanation made a previous consequence now invalid.
\end{example}

From this core %
notion of $\entails$, relating explanations to %
input-output, we can derive two ``homogeneous'' notions of ``entailment'', that is, from sequences of elements of a set to elements of the same set. \gppnew{This makes that notion more analogous to the notion of entailment in logic, which is defined from sets of formulas to a single formula.} One such notion is at input-output level, and the other at explanation level. For the former, we say $\seqio \deriveio (x,y)$ iff $\explainer(\seqio) \entails (x,y)$. For the latter, $\seq{\explanation} \deriveexp \explanation$ iff $\forall (x,y) \in \inputset \times \outputset$, if $\explanation \models (x,y)$ then $\seq{\explanation} \models (x,y)$ (summarised in Fig.~\ref{fig:diagram}).

\section{Consistency and non-monotonicity}

We now turn to properties of explainers, expressed by properties of \emph{consistency} and \emph{non-monotonicity} for the %
relations associated with explainers.

\begin{definition}[Consistency]
   A sequence of explanations $(\explanation_i)_{i \in [n]}$ is said to be \emph{consistent} iff there does not exist $x \in \inputset$, $y, y' \in \outputset$, with $y \neq y'$, such that $(\explanation_i)_{i \in [n]} \entails (x, y)$ and $(\explanation_i)_{i \in [n]} \entails (x, y')$.
   An entailment relation $\entails$ is said to be \emph{consistent} iff every sequence of explanations is consistent.
   A relation \deriveio{} is said to be \emph{consistent} iff there does not exist $x \in \inputset$, $y, y' \in \outputset$, with $y \neq y'$, and $((x_i,y_i))_{i \in [n]}$ such that $((x_i,y_i))_{i \in [n]} \entails (x, y)$ and $((x_i,y_i))_{i \in [n]} \entails (x, y')$.
\end{definition}

Since the relations $\deriveio$ and $\deriveexp$, derived from the base notion of $\entail$,
 are ``homogeneous''%
 , we can define %
 properties borrowed from the literature on non-monotonic reasoning \citep{generalpatterns,DBLP:journals/ai/KrausLM90}, what would not be possible for the relation $\entails$.
 We only generalise them to sequences, instead of sets (as typical)%
 . %

 Our hypothesis is that if they are important in (non-monotonic, defeasible) reasoning, and if the (interactive) explanation process can be seen as form of (non-monotonic) reasoning, then satisfying (or not) these properties may play a role on how users interact with explanations and what expectations they may (legitimately) have from them. \gppnew{Indeed, conditional reasoning has been explicitly evaluated in humans via questions motivated by such properties and, preliminary, support has been found for some properties, such as cautious monotonicity \citep{Neves2004AnET}, although the general picture is less clear in evaluations comparing human inferences with different non-monotonic logic formalisms \citep{DBLP:conf/ijcai/RagniEK16,Ragni2016FormalNT}. What seems, however, to be clear is how monotonic logic is insufficient for capturing conditional reasoning in humans \citep{byrne1989}.}

 Formally, some properties are:

\begin{definition}[Non-monotonicity] The relation $\entails$ is said to be \emph{non-monotonic} iff there is $(E_i)_{i \in [n]}$, $E_{n+1}$ and $(x,y)$ such that $(E_i)_{i \in [n]} \entails (x,y)$ and $(E_i)_{i \in [n+1]} \not\entails{} (x,y)$. Also, given a set $\setgen$, a relation $\derivegen$ from %
  $\seqset{\setgen}$ to $\setgen$, and $s, s_i \in \setgen$, for $i \in \naturalnumb$, the relation \derivegen{} is said to satisfy:%

\begin{itemize}
    \item \emph{non-monotonicity} iff there is
      $%
      (s_i)_{i \in [n]}, s_{n+1}, s$ %
      s.t. $(s_i)_{i\in [n]} \!\derivegen \!s$ and $(s_i)_{i \in [n+1]} \!\not\derivegen \!s$;
\item
\emph{reflexivity} iff for every $(s_i)_{i \in [n]}$ and $i$, $(s_i)_{i \in [n]} \derivegen{} s_i$;
\item
\emph{cautious monotonicity} iff for every $(s_i)_{i \in [n]}$, $s_{n+1},$ and $s$, if $(s_i)_{i \in [n]} \derivegen{} s_{n+1}$ and $(s_i)_{i \in [n]} \derivegen{} s$, then $(s_i)_{i \in [n+1]} \derivegen{} s$;
\item 
\emph{cut} iff for every $(s_i)_{i \in [n]}, s_{n+1}, s$, if $(s_i)_{i \in [n+1]} \derivegen{} s$ and $(s_i)_{i \in [n]} \derivegen{} s_{n+1}$, then $(s_i)_{i \in [n]} \derivegen{} s$.
\end{itemize}
\end{definition}

\gppnew{We can make some general considerations regarding those properties.
Reflexivity asserts that whatever is a premise should also be entailed. We would expect this to hold for the relation $\deriveio$ for many explainers, since violating this means that, after some interactions, an input-output pair could become unexplained by the current understanding of the sequence of explanations. That is, the interactive process fails for some asked input and returned output, which seems a failure of explanation. On the other hand, for $\deriveexp$, a failure of reflexivity would not be surprising. Indeed, Example~\ref{ex:nonmono} shows such a failure, where $(\explanation_0, \explanation_1) \not\deriveexp \explanation_0$.}

\gppnew{Regarding }cautious monotonicity, intuitively it captures the idea that if an expected behaviour is confirmed and made explicit, then other previously expected behaviours are still expected, i.e., a break of expectation is only caused by explicitly observing an unexpected behaviour).
\gppnew{For \deriveio{} this means that, when querying for an input which is classified as the (then) expected output, the resulting sequence of explanations, regardless of what it is, does not entail any less than what was entailed before. Cut, on the other hand, is in a way the converse: the resulting sequence of explanations does not entail any more than what was already entailed.}
\gppnew{As for \deriveexp{}, what cautious monotonicity means is that, if a sequence of explanations entails a single explanation, and this is appended to the sequence, then no entailment is lost. Again, cut means that no conclusion is gained.}

  \gppnew{We hypothesise that a violation of cut would be less severe than one of cautious monotonicity. Cut fails in some probabilistic logics \citep{generalpatterns}, and indeed one can think of further explanations as detailing the previous ones, even if already entailed, so that more cases are covered. Cautious monotonicity also has a pragmatical motivation of reducing updates in beliefs \citep[p.~12]{DBLP:journals/ai/KrausLM90}, which arguably could be relevant for the plausibility to, or comfort of a user in an interactive explanation process.}
  Which concrete methods satisfy which of those properties and what is the impact on user experience are open questions.

\begin{example}
  \label{ex:noncautiousmono}
  As a variation of Example~\ref{ex:nonmono}, assume the specificity interpretation and that for the input $x_1 = (20, 5)$, the output was $y=0$ and the second explanation was instead $\explanation_1 = \{%
  f > 10 \land %
  g \leq 0 \rightarrow 0, %
  f > 10 \land %
  g > 0 \rightarrow y = 1\}$.
  It is the case that $(\explanation_0) \entail (x_1, 0)$, and so $((x_0, 0)) \deriveio (x_1,0)$. Now consider a third input $x_2 = (20, -10)$. While $(\explanation_0) \entail (x_2, 0)$ (thus $((x_0, 0)) \deriveio (x_2,0)$), we now have that $(\explanation_0, \explanation_1) \entail (x_2, 1)$, since both rules in $\explanation_1$ are more specific than the one in $\explanation_0$. Therefore $((x_0, 0), (x_1, 0)) \deriveio (x_2,0)$, violating cautious monotonicity.
\end{example}

\subsection{Specificity for non-monotonic reasoning}
\gppnew{We will generalise here how the idea of specificity used in Examples~\ref{ex:nonmono} and \ref{ex:noncautiousmono} can be defined for general entailment relations $\entails$, or even to characterise one, in an abstract way.}

\begin{definition}
  A sequence of explanations $(\explanation_i)_{i \in [n]}$ is said to \textbf{cover} an input $x \in \inputset$ iff there is $y \in \outputset$ such that $(\explanation_i)_{i \in [n]} \models (x, y)$.

  A sequence of explanations $(\explanation_i)_{i \in [n]}$ is said to be \textbf{more specific than} another $\seq[m]{\explanation'}$ iff for every input $x \in \inputset$ that $\seq{\explanation}$ covers is also covered by $\seq[m]{\explanation'}$.
\end{definition}
In the definitions above, we can use single explanations as sequence of explanations, following our mentioned abuse of notation. We can now define what respecting specificity means for an entailment relation \entails{}.
\begin{definition}
  An entailment relation $\entails$ \emph{respects specificity} iff it is consistent and if $\explanation_{n+1}$ is more specific than $\seq{\explanation}$ and $\explanation_{n+1} \entails (x,y)$, then $\seq[n+1]{\explanation} \entails (x,y)$.
\end{definition}
A sceptical definition of $\models$ can be defined by extending the entailment relation from single explanations to sequences recursively via specificity. This exemplifies how one could define entailment in sequences from entailment in single explanations, based only on the abstract concept of entailment presented in this formalism, independent of instantiation.
The definition below captures such entailment relations, presenting the inductive step:
\begin{definition}
  An entailment relation $\entails$ is \emph{most sceptically specific} iff
  it is consistent, it respects specificity, and,
  if $\explanation_{n+1}$ is not more specific than $\seq{\explanation}$, for every $x \in \inputset$, $y,y' \in \outputset$:
  a) if $\explanation_{n+1} \entails (x,y)$ and $\seq{\explanation} \entails (x,y')$, with $y \neq y'$, then there is no $y'' \in \outputset$ such that $\seq[n+1]{\explanation} \entails (x,y'')$; else
  b) if $\explanation_{n+1} \entails (x,y)$ or $\seq{\explanation} \entails (x,y)$, then $\seq[n+1]{\explanation} \entails (x,y)$.
\end{definition}
\gppnew{This does not imply our framework is restricted to such entailment relations, as it can capture many other explanations, but we deem this special case to be of interest.}

\section{Discussion and future steps}
Consistency properties have already been advocated in the literature, as discussed in the introduction. The link between classification itself and NMR is less studied, but NMR-inspired properties have been presented for the task of classification and applied for a specific method built on argumentation \citep{DBLP:conf/kr/Paulino-PassosT21}. A related but different direction is using interactions and explanations for changing the model itself or predictions for a particular instance \citep{DBLP:journals/ai/SreedharanCK21,DBLP:journals/ai/RagoCBLT21}.
\citet{DBLP:conf/ecsqaru/Amgoud21} also presents explanations as functions that can be non-monotonic, showing examples on how an explanation can be non-monotonic and proposing an argumentation-based approach to create non-monotonic explanations without losing consistency. Our main differences from this work are: i) we base our definitions on sequences, which are more general and can capture the interactive scenario; ii) we make explicit the entailment relations, linking explicitly to NMR (at both input-output and explanation levels).

\gpp[inline]{a further discussion not included: \\ (improve and add in the full paper:) The argumentation-based proposal is limited, in that the semantics used consider maximal sets of arguments which do not attack each other (conflict-free). Indeed, this semantics does not use the direction of attack at all. This does not incorporate more complex way of considering the arguments, such as how arguments may be stronger than others. On the other hand, our proposed specificity criterion is an example on how this could happen.}

A line of work with connections to this as well is on building model interpretation (global explanations) from local explanations \citep{DBLP:journals/ai/SetzuGMTPG21,DBLP:journals/corr/abs-2205-00130}. Indeed, one possible instantiation of $\entail$ is by generating local explanations $\seq{\explanation}$ for $\seqio$, and then aggregating the resulting explanations into a single, global, explanation $\explanation_{agg}$, and saying $\seq{\explanation} \models (x,y)$ exactly when $\explanation_{agg} \models (x,y)$

In summary, we presented a first approach to modelling an interactive explanation scenario, and discussed possible properties of it, specially ones based on non-monotonic reasoning (NMR). We also show how the general concept of specificity in NMR may be used to extend entailment from single explanations to sequences of them. We believe considerations inspired by non-monotonic reasoning can inspire questions and angles of study of explainability methods. We leave to future work analyses of specific systems and algorithms, as well as empirical evaluations of the impact of the presence or absence of particular NMR properties with human users.

\section*{Acknowledgments}
The first author was supported by Capes (Brazil, Ph.D. Scholarship 88881.174481/2018-01).
The second author was partially funded by the European Research Council (ERC) under the European Union’s Horizon 2020 research and innovation programme (grant agreement No.
101020934) and by J.P.
Morgan and by the Royal Academy of Engineering under the Research Chairs and Senior Research Fellowships
scheme. Any views or opinions expressed herein are solely
those of the authors.

\bibliographystyle{named}
\bibliography{references}

\end{document}